\documentclass{article}

\input{./preamble/preamble.tex}

\usepackage{centernot}
\usepackage{amsthm}

\crefname{lemma}{lemma}{lemmas}
\Crefname{lemma}{Lemma}{Lemmas}
\crefname{thm}{theorem}{theorems}
\Crefname{thm}{Theorem}{Theorems}
\crefname{prop}{proposition}{propositions}
\Crefname{prop}{Proposition}{Propositions}

\newtheorem{lemma}{Lemma}

\newcommand{\cN}{\mathcal{N}}

\newcommand{\g}{\, | \,}

\usepackage{booktabs,arydshln}
\makeatletter
\def\adl@drawiv#1#2#3{%
        \hskip.5\tabcolsep
        \xleaders#3{#2.5\@tempdimb #1{1}#2.5\@tempdimb}%
                #2\z@ plus1fil minus1fil\relax
        \hskip.5\tabcolsep}
\newcommand{\cdashlinelr}[1]{%
  \noalign{\vskip\aboverulesep
           \global\let\@dashdrawstore\adl@draw
           \global\let\adl@draw\adl@drawiv}
  \cdashline{#1}
  \noalign{\global\let\adl@draw\@dashdrawstore
           \vskip\belowrulesep}}
\makeatother

\newcommand{\bas}[1]{\begin{align*}#1\end{align*}}

\newcommand{\ba}[1]{\begin{align}#1\end{align}}

\newcommand{\distas}[1]{\mathbin{\overset{#1}{\kern\z@\sim}}}

\newcommand{\uv}{\boldsymbol{u}}

\newcommand{\zv}{\boldsymbol{z}}

\newcommand{\phiv}{\boldsymbol{\phi}}

\newcommand{\bE}{\mathbb{E}}

\usepackage{amsmath}
\usepackage{subcaption}
\usepackage{placeins}

\usepackage{wrapfig}

\usepackage{dsfont}
\usepackage{tcolorbox}

\DeclareMathOperator*{\argmax}{arg\,max}

\usepackage[nameinlink]{cleveref}
\crefname{figure}{fig.}{figs.}
\Crefname{figure}{Fig.}{Figs.}
\crefname{equation}{eq.}{eqs.}
\Crefname{equation}{Eq.}{Eqs.}
\crefname{algorithm}{alg.}{algs.}
\Crefname{algorithm}{Alg.}{Algs.}
\crefname{section}{\S}{\S\S}
\Crefname{section}{\S}{\S\S}
\crefname{lemma}{lemma}{lemmas}
\Crefname{lemma}{Lemma}{Lemmas}
\crefname{theorem}{theorem}{theorems}
\Crefname{theorem}{Theorem}{Theorems}
\crefname{proposition}{proposition}{propositions}
\Crefname{proposition}{Proposition}{Propositions}

\usepackage{authblk}

\author[1]{ Mingzhang Yin}
\author[2]{ Ruijiang Gao}
\author[1]{ Weiran Lin}
\author[1]{ Steven M. Shugan}

\affil[1]{ \small University of Florida, Warrington College of Business}
\affil[2]{ \small University of Texas at Austin, McCombs School of Business}

\begin{document}

\title{Nonparametric Discrete Choice Experiments with \\ Machine Learning Guided Adaptive Design}

\maketitle

\begin{abstract}
Designing products to meet consumers' preferences is essential for a business's success. 
We propose Gradient-based Survey (GBS), a discrete
choice experiment for multiattribute product design. 
The experiment elicits consumer preferences through a sequence of paired comparisons for partial profiles. GBS adaptively constructs paired comparison questions based on the respondents' previous choices. Unlike the traditional random utility maximization paradigm, GBS is robust to model misspecification by not requiring a parametric utility model. Cross-pollinating the machine learning and experiment design, GBS is scalable to products with hundreds of attributes and can design personalized products for heterogeneous consumers. We demonstrate the advantage of GBS in accuracy and sample efficiency compared to the existing parametric and nonparametric methods in simulations. ~\looseness=-1
  
\end{abstract}

\section{Introduction}

Identifying an optimal product based on consumers' preferences is essential for the success of a business. Such a problem is prevalent in companies where the product consists of multiple attributes such as health insurance,
cell phone plans, pizzas, automobiles, logos, and email advertisements \citep{balakrishnan2004development,netzer2011adaptive,bertsimas2017robust,Ellickson2022-uo}. In this paper, we focus on products with discrete attributes represented as multivariate binary variables, as is the case in A/B testing \footnote{A discrete attribute with more than two levels can be coded as multiple binary variables.}. \looseness=-1 

Selecting appropriate products from a choice set is complicated for decision-makers. First, the desired product should meet the unobserved consumer preferences. The latent preferences are often revealed by survey experiments known as conjoint analysis \citep{Luce1964-ge,green1971conjoint}. Since the milestone work of \cite{louviere1983design}, the choice-based conjoint (CBC) analysis become one of the most widely used methods to quantify multiattribute preference \citep{hein2020analyzing}. A prevalent assumption is that the preference for an attribute is quantified by a part-worth score, and the total utility of a product profile is the sum of part-worths \citep{green1978conjoint}. This parametric assumption, however, may oversimplify how respondents encode and evaluate products \citep{Allenby2005}. Second,
the choice set for product design grows exponentially with the number of attributes, making the problem NP-hard \citep{kohli1989optimal}. The problem is more evident with the development of technology when more and more components are integrated into a single product. For example, the design of a smartphone might need to consider hundreds of attributes from a digital camera, screen display, connectivity modules, software applications, and a range of sensors. The high-dimensional attributes pose a scalability challenge to the extant product design methods. Lastly, consumer preference is heterogeneous, and it is desired to provide a customized product aligned with individual tastes. Nowadays, an increasing number of products are presented as digital content, making personalized product design feasible. For example, the email campaign content can be designed based on a receiver's demographic information. The fast development of generative AI might expand the need for personalized product design. 

\begin{figure}
  \centering
  \includegraphics[width=.5\linewidth]{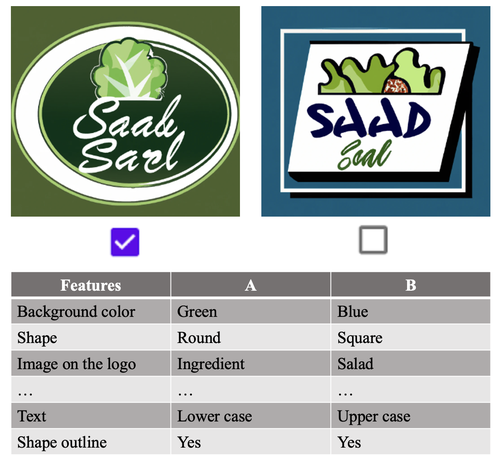}
  \caption{Demonstration of a paired choice question for a logo design.}
  \label{fig:demo}
\end{figure}

We propose GBS that is robust to model misspecification of the utility function, scalable to high-dimensional attributes and applies to single or personalized product design. The idea of GBS is to combine gradient-based machine learning with discrete choice experiments (DCEs). We model the product attributes as random variables following Bernoulli distributions, then maximize an objective function such as the market share. The inference of the optimal product is conducted by computing the gradient of the objective function with respect to the Bernoulli distribution parameters. However, it is challenging to compute the gradient because the genuine functional form of respondent choice given an item is not explicit and the gradient cannot be directly computed with non-continuous variables. To address these challenges, we adopt the score function method, a.k.a. the policy gradient method in reinforcement learning, to compute the gradient using data from DCEs. We build on the recent development of discrete optimization in machine learning to develop new variance reduction tools for the score function gradient. The unbiased and low-variance gradient is then used to generate the next survey questionnaire and update the optimal product's distribution parameter by stochastic gradient descent (SGD).  \looseness=-1

Bridging gradient-based machine learning and experiments borrows strength from both worlds. The sequential nature of SGD naturally provides an adaptive approach to design experiments. Unlike traditional heuristic or rule-based adaptive design \citep{green1991adaptive,netzer2011adaptive}, the gradient method is derived mathematically and is readily to incorporate new statistical tools for variance control. The proposed GBS maximizes the information extracted from each paired comparison question by utilizing the greedy property and the variance reduction technique of the gradient. GBS collects the data in the same way as a classic paired conjoint design where the respondents are randomly selected from a population and asked to choose between a pair of items with different attributes. Hence, GBS can be seamlessly integrated into commercial adaptive conjoint software like Sawtooth Software \citep{huber2005conjoint}. On the other hand, the experiments 
generate the data in a way by which machine learning can explore combinatorial actions. In contrast, using observational data may face the overlapping and extrapolation problems because the data in practice may only be collected on a small subset of attribute combinations from the large action space. 

In simulation, we demonstrate that GBS scales to hundreds of attributes efficiently and can infer the optimal products accurately. We compare GBS with parametric and nonparametric baseline methods. Over a variety of utility functions, GBS is more robust in model specification than parametric methods and is more sample-efficient than neural networks. Finally, we apply GBS to learn an individual policy with combinatorial actions via experiments.

\section{Problem Set Up}

Denote $Y_i(Z, Z_0)$ as the user $i$'s choice of product given the focal product $Z$ and a baseline product $Z_0$. The products are represented by $K$ binary features $Z, Z_0 \in \{0,1\}^K$. The baseline product $Z_0$ can be a competitor's product, the exisitng product in the market, or an empty set. The potential outcome $Y_i(Z, Z_0)$ is distributed according to the unknown data generating process. $Y_i(Z_1, Z_0)=1$ if user $i$ chooses $Z_1$ and $Y_i(Z_1, Z_0)=0$ if $Z_0$ is chosen. ~\looseness =-1

We first consider how to identify a single optimal product. The objective is 
\ba{
\max_Z V(Z) = \bE[Y_i(Z, Z_0)],
\label{eq:single}
}
which represents the potential market share of the proposed product $Z$ in the presence of the baseline product. The difficulties in solving \Cref{eq:single} are that the number of possible products grows exponentially with feature numbers, and the functional form of $Y_i(Z, Z_0)$ is unknown and might be complicated.

\section{Gradient-based Survey Design}
\label{sec:gradient}
We adopt the Random Utility Maximisation (RUM) framework to model the choice behavior \citep{mcfadden1974measurement}. Each product $Z_j$ is associated with a utility $U_i(Z_j)$ for individual $i$. The alternative $Z_1$ is chosen from the pair $(Z_1, Z_0)$ if and only if $U_i(Z_1) > U_i(Z_0)$. In RUM, the utility is decomposed as $U_i(Z_j) = V_i(Z_j) + \epsilon_{ij}, j=0,1$ where $V_i(Z_j)$ is often called the representative
utility \citep{train2009discrete}. We assume the random $\epsilon_{ij}$ independently follows type I extreme value (Gumbel) distribution. %
Accordingly, the probability of choosing item $Z$ is $p(Y_i(Z, Z_0) = 1) =  \exp(V_i(Z)) / (\exp(V_i(Z)) + \exp(V_i(Z_0)))$. Notice that the product $Z$ that maximizes the choice probability $p(Y_i(Z, Z_0) = 1)$ is irrelevant to what the baseline product $Z_0$ is. We do not make a parametric assumption for the representative utility $V_i(Z)$.

Instead of optimizing the discrete features $Z$ directly, we consider $Z$ as random variables following distribution $p(Z;\pi) = \prod_{k=1}^K\text{Bern}(z_k; \pi_k)$, $\pi_k \in[0,1]$. We then transform the problem in \Cref{eq:single} to an equivalent problem with the same optimal solution as
\ba{
  \max_\pi  V(\pi) = \bE_{Z \sim p(Z;\pi)}\bE[Y_i(Z, Z_0)\g Z]. 
  \label{eq:single2}
}
The equivalence of probabilistic reformulation is shown in \cite{yin2020probabilistic} Theorem 1. To facilitate the optimization in an unconstrained space, we parameterize the probability by the sigmoid function $Z \sim \prod_{k=1}^K \text{Bern}(z;\pi_k = \sigma(\phi_k)), \phi \in \mathbb{R}^K$, $\sigma(x) = 1/(1+e^{-x})$, and optimize $V(\phi):=V(\pi = \sigma(\phi))$ with the logits $\phi$. 

The gradient of \Cref{eq:single2} can be computed by the score function estimator (a.k.a. REINFORCE) as $\nabla_\phi V(\phi) = \bE_{Z \sim p(Z;\phi)} [\nabla_{\theta} \log p(Z;\phi) \bE[Y_i(Z, Z_0)\g Z] ]$. However, score function gradients often suffer from high variance, and many works have been devoted to reducing the variance. With a direct application of the antithetic sampling and control variates \citep{yin2019arm} to the product design problem, we have the following Lemma. 

\begin{lemma} \label{lemma:grad}
  An unbiased gradient of the objective $V(\phi)$ is
  \ba{
  \nabla_\phi V(\phi) = \bE_{u \sim \prod_{k=1}^K\text{Unif}(0,1)} \Big[\bE\big[(Y_i(Z_1(u),Z_0) - Y_i(Z_2(u),Z_0))(u-\frac12) \g u \big]\Big],
  \label{eq:grad}
  }
  where $Z_1(u) = \mathbf{1}[u > \sigma(-\phi)], Z_2(u) = \mathbf{1}[u < \sigma(\phi)].$
\end{lemma}
For completeness, the proof is in the Appendix. The gradient in \Cref{eq:grad} is guaranteed to reduce the variance of score function gradient for non-negative objective \citep{yin2019arm}. A Monte-Carlo estimate of the gradient in \Cref{eq:grad} can be computed by asking respondent $i$ to choose between products $Z_1$ and $Z_0$,  between $Z_2$ and $Z_0$, and take the difference of the choices. However, such question design might be sensitive to the selection of the baseline $Z_0$. A strong or weak $Z_0$ may make the choices $Y_i(Z_1,Z_0) $ and $Y_i(Z_2,Z_0)$ the same frequently, resuting in a zero gradient and slow learning. Moreover, a respondent's preferences across two choices may not be fully consistent and comparable.

To mitigate these problems, we marginalize out the zero gradients using \Cref{lemma:zero}. 

\begin{lemma}\label{lemma:zero}
$Y(Z_1,Z_0) - Y(Z_2,Z_0) \g Y(Z_1,Z_0) \neq Y(Z_2,Z_0) \stackrel{d}{=} (2Y(Z_1, Z_2) - 1)$
\end{lemma}

With \Cref{lemma:zero,lemma:grad}, we have the following result.

\begin{lemma}\label{lemma3}
  $\tilde{g} = (2Y_i(Z_1(u), Z_2(u)) - 1) (u-\frac12) p(\mathcal{A}_i)$ satisfies $\bE[\tilde{g}] = \nabla_\phi V(\phi)$, where $Z_1(u)$, $Z_2(u)$ are defined as in \Cref{lemma:grad},$u \sim \prod_{k=1}^K\text{Unif}(0,1)$, event $\mathcal{A}_i = \{Y_i(Z_1,Z_0) \neq Y_i(Z_2,Z_0)\}$.
\end{lemma}
Computing $\tilde{g}$ takes a choice  $Y_i(Z_1(u), Z_2(u))$ from a random respondent $i$ between two partial profiles $Z_1(u)$ and $Z_2(u)$. The baseline product $Z_0$ is no longer needed. %
The partial profile comparison alleviates cognitive burden and elicits respondents' preferences more accurately than a choice from a large action space (a demonstration is in \Cref{fig:demo}). $Z_1(u)$ and $Z_2(u)$ differ in feature $k$ with probability $1-|2\pi_k-1|$, which increases monotonically from 0 to 1 with the variance $\pi_k(1-\pi_k)$. It is aligned with the intuition that in order to maximize the information gain from each question, features with high uncertainty should have a high chance of being asked. 
 Though the quantity $p(\mathcal{A}_i)$ is unknown, %
it is a constant shared by all the elements of the stochastic gradient $\tilde{g}$. Therefore, the constant can be absorbed in the stepsize of the SGD and does not affect the convergence. Finally, we propose the gradient estimate for GBS with $Z_1(u), Z_2(u)$ defined in \Cref{lemma:grad} as
\ba{
g_{\text{GBS}} = (2Y_i(Z_1(u), Z_2(u)) - 1) (u-\frac12), ~~~u \sim \prod_{k=1}^K\text{Unif}(0,1).
\label{eq:gbs}
}
The steps of GBS are summarized in \Cref{alg:gbs}.

The data collection of GBS shares the same form of survey with the paired CBC, which has been widely applied to understanding individual preferences in marketing, politics, and other computational social science for decades \citep{Luce1964-ge,toubia2003fast,Hainmueller2015-uy,Egami2019-mm,Goplerud2022-zq}.  GBS is thus compatible with many existing survey systems. 
The paired choice design used in practice requires a single choice for each question and rules out the situation of choosing both or none, which echos the step of  zero gradients marginalization in \Cref{eq:grad}.  In a nutshell, GBS uses the gradient information to automatically and adaptively design experiments, and in the meanwhile, uses the data from the experiments to estimate a low-variance stochastic gradient for the optimal product identification. ~\looseness = -1

\begin{algorithm}[t]
  \small{
  \SetKwData{Left}{left}\SetKwData{This}{this}\SetKwData{Up}{up}
  \SetKwFunction{Union}{Union}\SetKwFunction{FindCompress}{FindCompress}
  \SetKwInOut{Input}{input}\SetKwInOut{Output}{output}
  \Input{Number of features $K$, number of questions per respondent $n_q$, stepsize $\eta$
  }
  
  Initialize the logits $\phi$ randomly
  
  \While{not converged}{

   Sample a random individual $i$ from the population.

   \For{$j = 1,2 \cdots n_q$}{
    Sample $u \sim \prod_{k=1}^K\text{Unif}(0,1)$

    Generate a pair of products $Z_1(u) = \mathbf{1}[u > \sigma(-\phi)], Z_2(u) = \mathbf{1}[u < \sigma(\phi)]$
    
    Record the respondent's choice $Y_i(Z_1(u), Z_2(u))$

    Compute the gradient estimate $g_{\text{GBS}} = (2Y_i(Z_1(u), Z_2(u)) - 1) (u-\frac12)$

    Update $\phi \leftarrow \phi + \eta g_{\text{GBS}}$
   }

  }
  \caption{Gradient-based Survey for Product Design}
  \label{alg:gbs}}
\end{algorithm}

\section{Individualized Policy Learning}

A single product is often not optimal for all users. For example, a personalized advertisement email designed based on individual shopping history may improve the consumers' open rate. We consider the problem of learning an individualized policy that assigns a customized product to each user. %

Suppose the covariates $X_i$ of individual $i$ are observed. The optimization objective is
\ba{
\max_{\theta} V(\theta) = \bE_{X_i \sim p(X)}\bE_{Z_i \sim p(Z; \pi_\theta(X_i))} \bE[Y_i(Z_i, Z_0)],
}
where $p(Z; \pi_\theta(X_i)) = \prod_{k=1}^K\text{Bern}(z_k; \pi_\theta(X_i)_k)$. The policy $\pi_\theta(X_i) = \sigma(\phi_i)$ and  the logits $\phi_i= g(X_i; \theta) \in \mathbb{R}^K$  are the output of an amortized neural network parameterized by $\theta$ with input $X_i$. 

Applying the results in \Cref{sec:gradient}, an unbiased Monte Carlo estimate of the gradient w.r.t. the logits $\phi_i$ is $c_i(2Y_i(Z_1(u), Z_2(u)) - 1) (u-\frac12)$ where  $u \sim \prod_{k=1}^K\text{Unif}(0,1)$ and $c_i$ is a scalar to be absorbed in the stepsize. Using the chain rule, the GBS gradient of policy parameter $\theta$ is ~\looseness=-1
\ba{
  g_{\text{GBS}}(\theta) = (2Y_i(Z_1(u), Z_2(u)) - 1) (u-\frac12)^\top  \frac{\partial g(X_i; \theta)}{\partial \theta}, \quad u \sim \prod_{k=1}^K\text{Unif}(0,1)
  \label{eq:grad4}
}
with $Z_1(u), Z_2(u)$ defined in \Cref{lemma:grad}. When the action space is combinatorial, existing policy learning methods using offline collected data often face the extrapolation problem because there is not enough variation in the observed sets of items \citep{sachdeva2020off}.  We are the first to enable policy learning with combinatorial actions using conjoint experiments. The algorithm is summarized in \Cref{alg:gbs-ind} in \Cref{app:add}. ~\looseness=-1

\section{Related Work}

\paragraph{Conjoint analysis.}
Conjoint analysis (CA) uses a survey-based experiment design to measure multidimensional preferences \citep{Luce1964-ge,toubia2003fast}. We collect the data with a paired comparison survey similar to CA. Different from CA, GBS does not assume a linear additive utility model and is applicable to cases with nonlinear interactive utilities. CA estimates partworths as the choice model parameters \citep{Hainmueller2015-uy} while GBS does not estimate a choice model and identifies the optimal product using the choice data directly.

\paragraph{Adaptive experimental design. }

Adaptive conjoint analysis (ACA) progressively refines the attribute levels presented to respondents for more accurate and efficient data collection \citep{toubia2004polyhedral}. For example, D-efficiency is designed to maximize Fisher information \citep{kuhfeld1994efficient}, polyhedral methods combine geometric intuition and analytic center technique to shrink feasible region \citep{toubia2003fast,toubia2007probabilistic,saure2019ellipsoidal}, and adaptive self-explication integrates attribute importance in design \citep{netzer2011adaptive}. It might be challenging to generalize heuristics to high-dimensional features.  %
GBS design is derived from variance reduction of score function gradient and is aligned with the uncertainty reduction intuitions. Apart from ACA, an active learning approach is proposed to learn nonparametric choice models using a directed acyclic graph.  Nodes in such a graph are alternative profiles, which limit the node number to a small scale \citep{susan2022active}. ~\looseness=-1 

\paragraph{Product design by optimization. }
Existing product design methods often adopt optimization heuristics like Genetic Algorithms \citep{balakrishnan1996genetic, balakrishnan2004development}, simulated annealing \citep{tsafarakis2016redesigning}, evolutionary algorithm, and beam
search  \citep{Paetz2021,hauser2011new}. Though achieving empirical improvements, the properties and generalization abilities of the heuristics is largely unclear. Discrete optimization methods such as Lagrangian relaxation with branch-and-bound have also been applied \citep{camm2006conjoint,belloni2008optimizing}. More broadly, deep learning methods such as variational autoencoders are used for the design of product aesthetics \citep{burnap2023product}. ~\looseness=-1

\paragraph{Policy learning. } The customized product design is related to policy learning. The offline policy learning often estimates the outcome function \citep{Wang2016-js} or directly optimizes the value function by propensity weighting \citep{Athey2017-hr}. However, it is hardly possible for the observational data to contain the outcomes of all the actions from a combinatorial space. The learned policy might be suboptimal due to the overlapping issue. Online policy learning uses bandits and reinforcement learning to maximize the cumulative reward. \citet{qin2014contextual} explores combinatorial action spaces, which require users to select actions from the complete action space and need external covariates for each feature dimension. GSB overcomes these challenges using the partial profile design from the conjoint analysis.

\section{Empirical Study}
\label{sec:exp}
We compare GBS with baseline models using simulated data similar to \citet{toubia2007probabilistic}. The product $Z$ is represented as $K$ binary features. First, we study the problem of identifying a single optimal product. For each independent trial, the population-level marginal preferences for the features are generated by $\mu \sim \cN(a, I)$, $a=(1,\cdots,1)$. The individual preferences (partworths) are generated by $W_i \sim \cN(\mu,I)$ for each individual $i$.

\paragraph{Data generation.} We consider three types of representative utilities $V_i(Z)$. The first type is linear utility $V_i(Z) = W_i^\top Z$. The second type includes pairwise interactions  $V_i(Z) = W_i^\top Z + \sum_{k,k'} \tilde{W}_{i}^{kk'} Z_k Z_{k'}$ where $\tilde{W}_{i}^{kk'} \sim \text{Unif}(-2a, 0)$. For $K=10$, all the pairwise interactions are 
included, and for $K=100$, a subset of 100 pairwise interaction terms is used to compute $V_i(Z)$. The third type set $V_i(Z) = f_0(Z)$ where $f_0(\cdot)$ is a pre-trained neural network. This type of utility includes higher-order interactions of the product features. The data for each method are the query product features and the choices collected from paired choice questions, i.e., $\{(Y_i(Z_{1}^{ij}, Z_{0}^{ij}), Z_{1}^{ij}, Z_{0}^{ij})\}_{i=1:N}^{j=1:n_q}$. For non-adaptive methods, the item pair in a question is generated randomly. Each respondent makes $n_q=10$ times the choices. 

\paragraph{Baselines.} A logistic model (Logistic) assumes $\hat{V}_i(Z) = W^\top Z$. For a pair of products $(Z_1, Z_2)$, the likelihood of choosing $Z_1$ is $1/(1+\exp(W^\top(Z_2-Z_1)))$. The parameter $W$ is estimated by maximum likelihood and the optimal product is $\mathbf{1}[\hat{W}_{MLE}>0]$. A mixed logit model is a hierarchical Bayes (HB) model widely used in conjoint analysis \citep{Allenby2005}. It assumes $m \sim \cN(0, I)$, $w_i \g m \sim \cN(\mu, I)$, and $p(Y(Z_1,Z_2) = 1 \g w_i, Z_1, Z_2) = 1 / (1+\exp(w_i^\top(Z_2-Z_1)))$. We estimate $m$ as the maximum a posteriori estimation using the PyMC package, and the estimated optimal product is $\mathbf{1}[\hat{m}_{MAP}>0]$. Another baseline takes the representative utility $\hat{V}_i(Z) = f_\gamma(Z)$. $f_\gamma(\cdot)$ is a feedforward neural network (NN) with two hidden layers and parameters $\gamma$. $\gamma$ is estimated as an MLE. The estimated optimal product is $\argmax_{Z} f_\gamma(Z)$, which requires enumerating all possible products. 

\paragraph{Evaluation metrics.} The first metric for evaluating a chosen product is the average utility on a hold-out test set. It compares the relative performance of different methods. When computationally feasible, we also rank all the possible products according to their average utility on the population from high to low. The ranking allows the evaluation of absolute performance compared to the global optimum. 

\begin{figure}[t]
  \centering
  \begin{subfigure}{.33\textwidth}
  \centering
  \includegraphics[width=.95\linewidth]{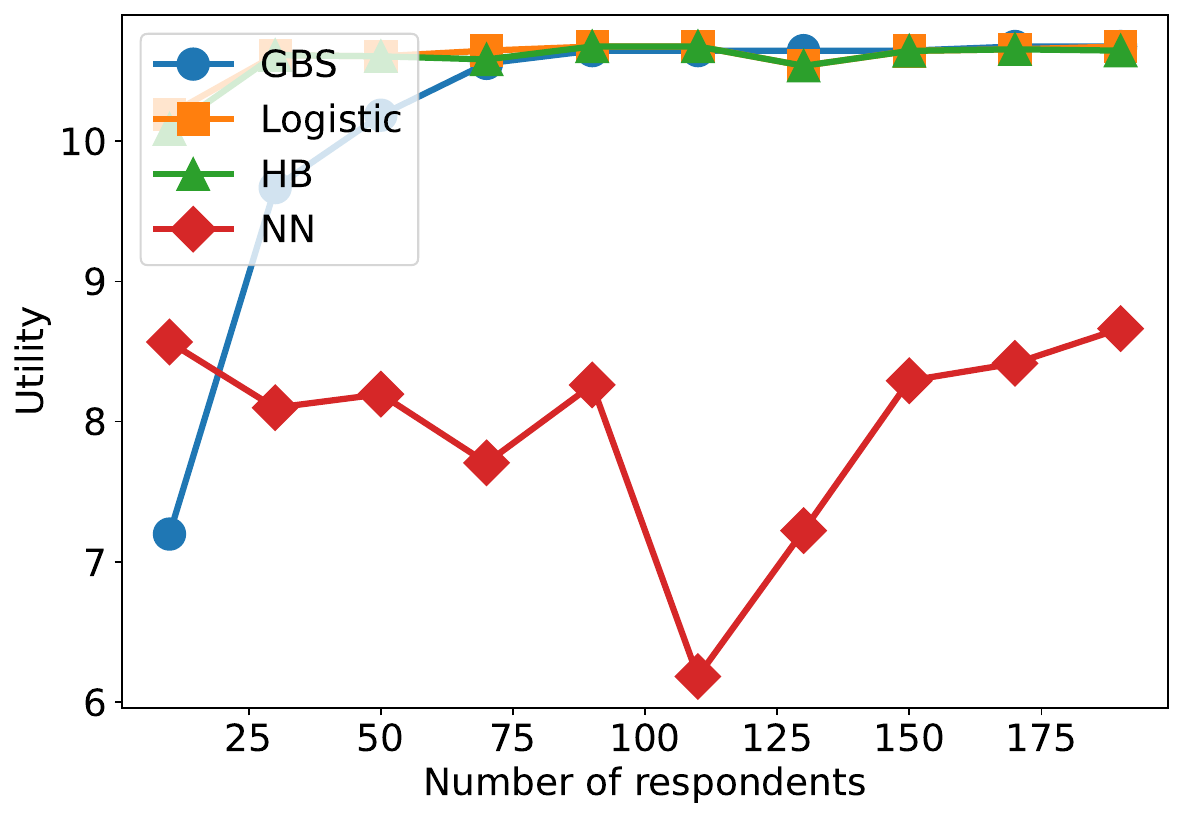}
  \end{subfigure}%
  \begin{subfigure}{.33\textwidth}
  \centering
  \includegraphics[width=.95\linewidth]{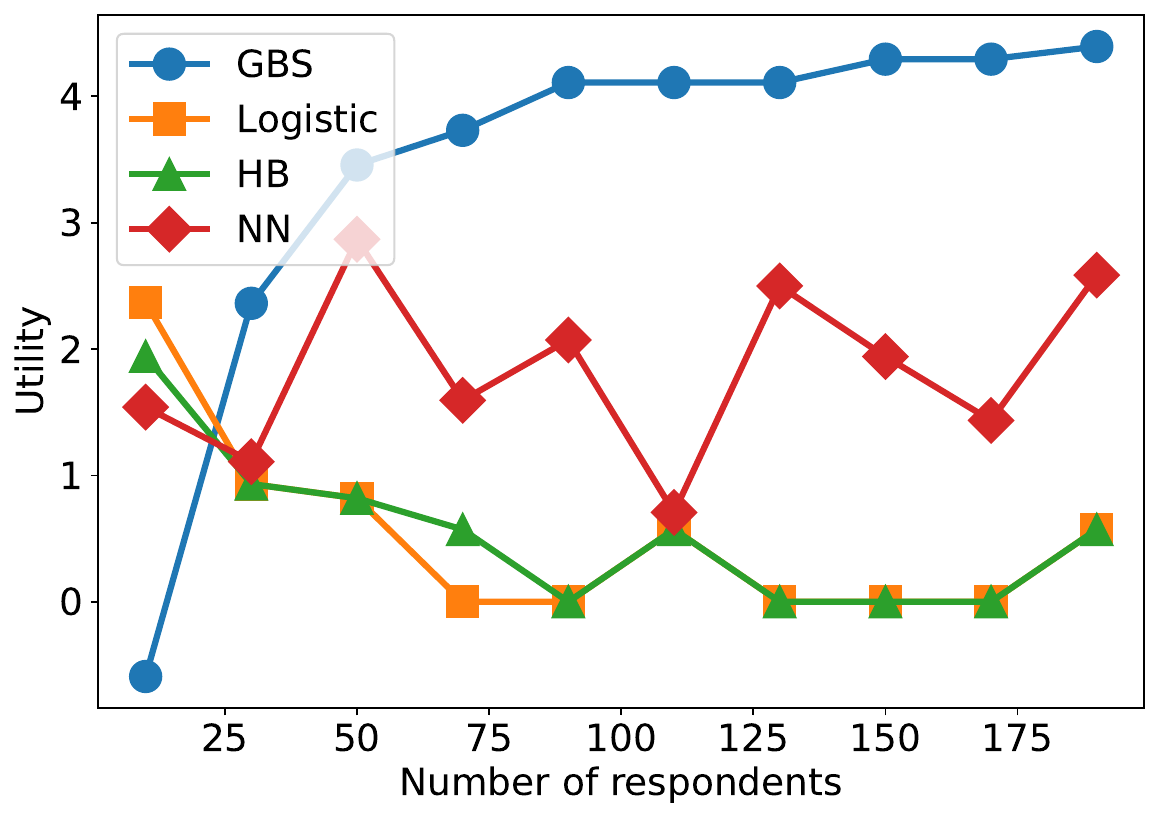}
  \end{subfigure}
  \begin{subfigure}{.33\textwidth}
    \centering
    \includegraphics[width=.95\linewidth]{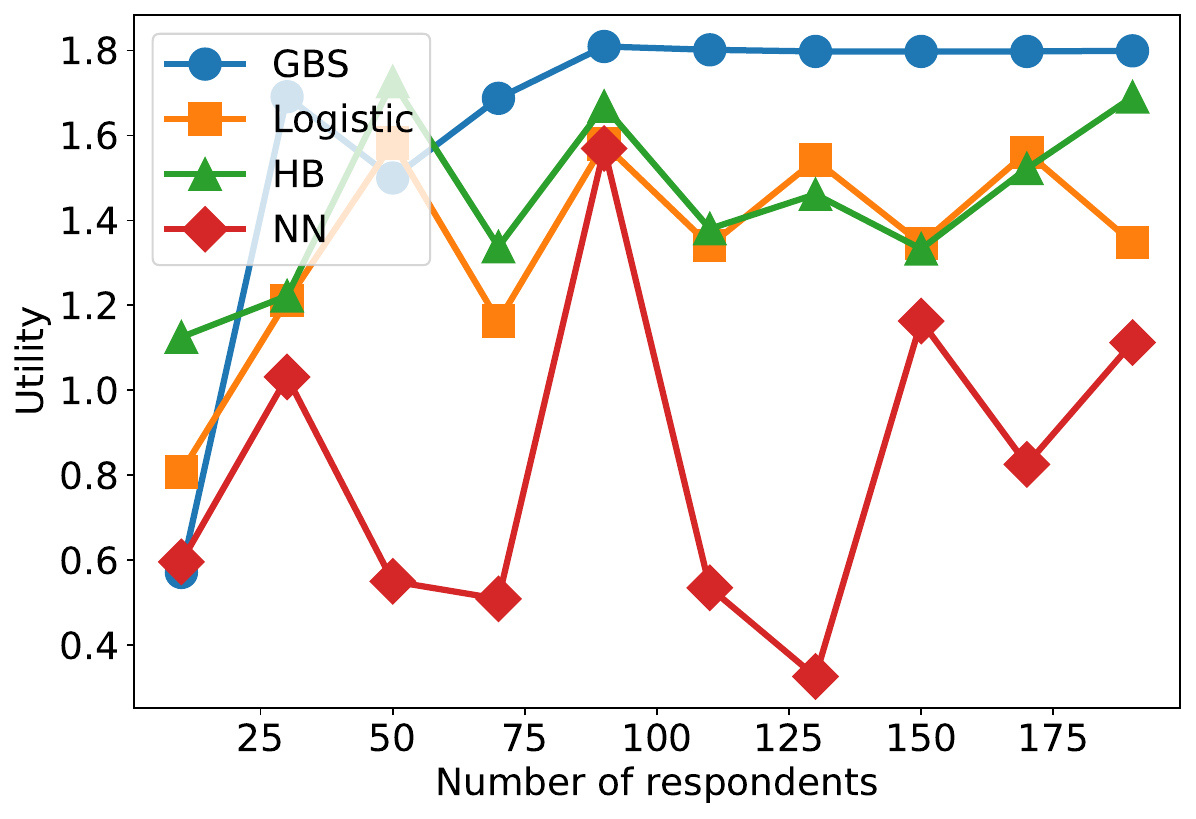}
    \end{subfigure} \\
    \begin{subfigure}{.33\textwidth}
      \centering
      \includegraphics[width=.95\linewidth]{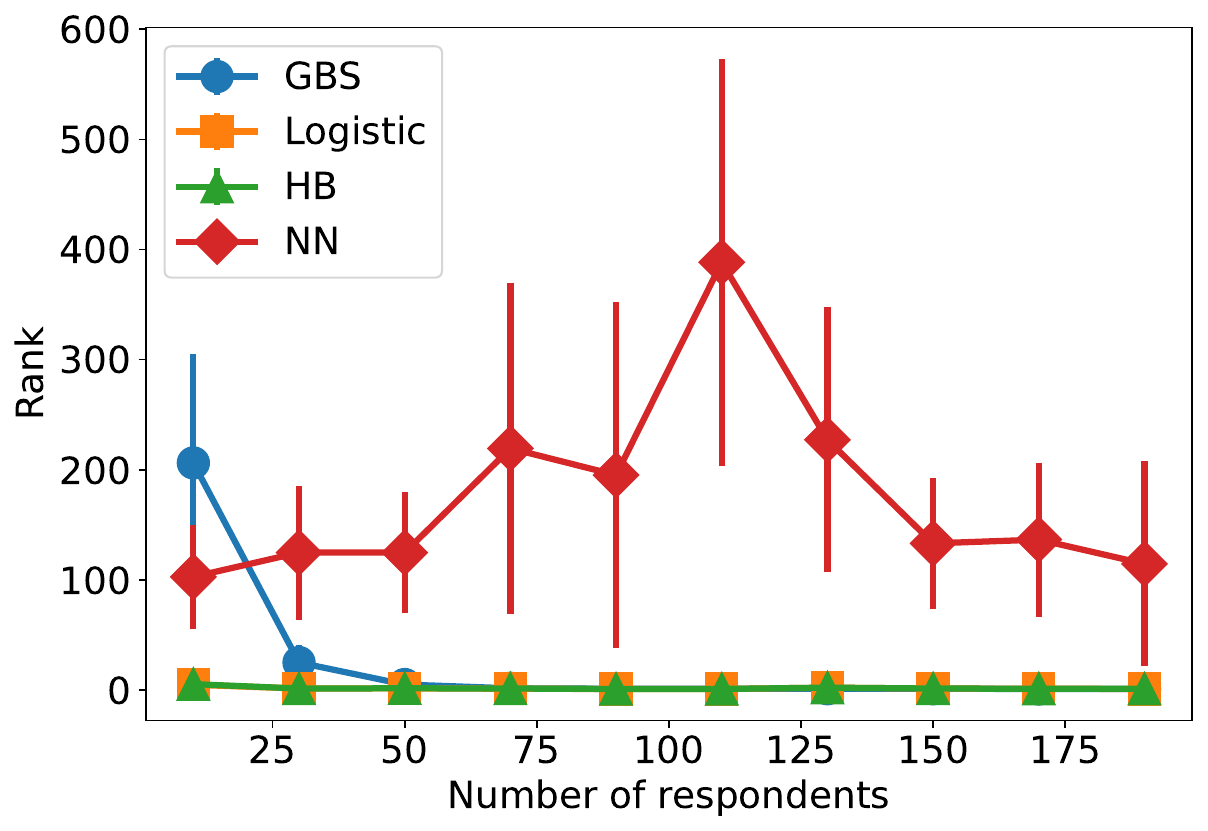}
      \caption{Type 1}
      \label{fig:rt1k10}
      \end{subfigure}%
      \begin{subfigure}{.33\textwidth}
      \centering
      \includegraphics[width=.95\linewidth]{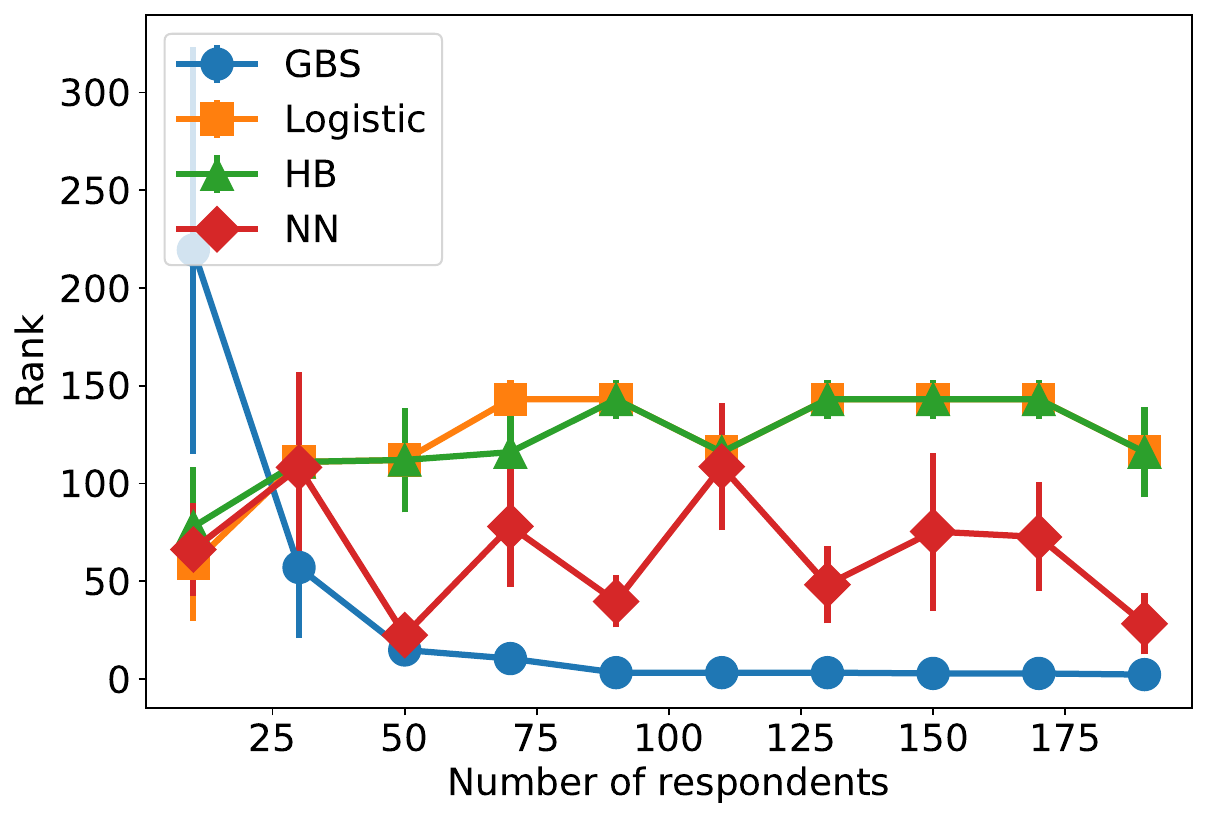}
      \caption{Type 2}
      \label{fig:rt2k10}
      \end{subfigure}
      \begin{subfigure}{.33\textwidth}
        \centering
        \includegraphics[width=.95\linewidth]{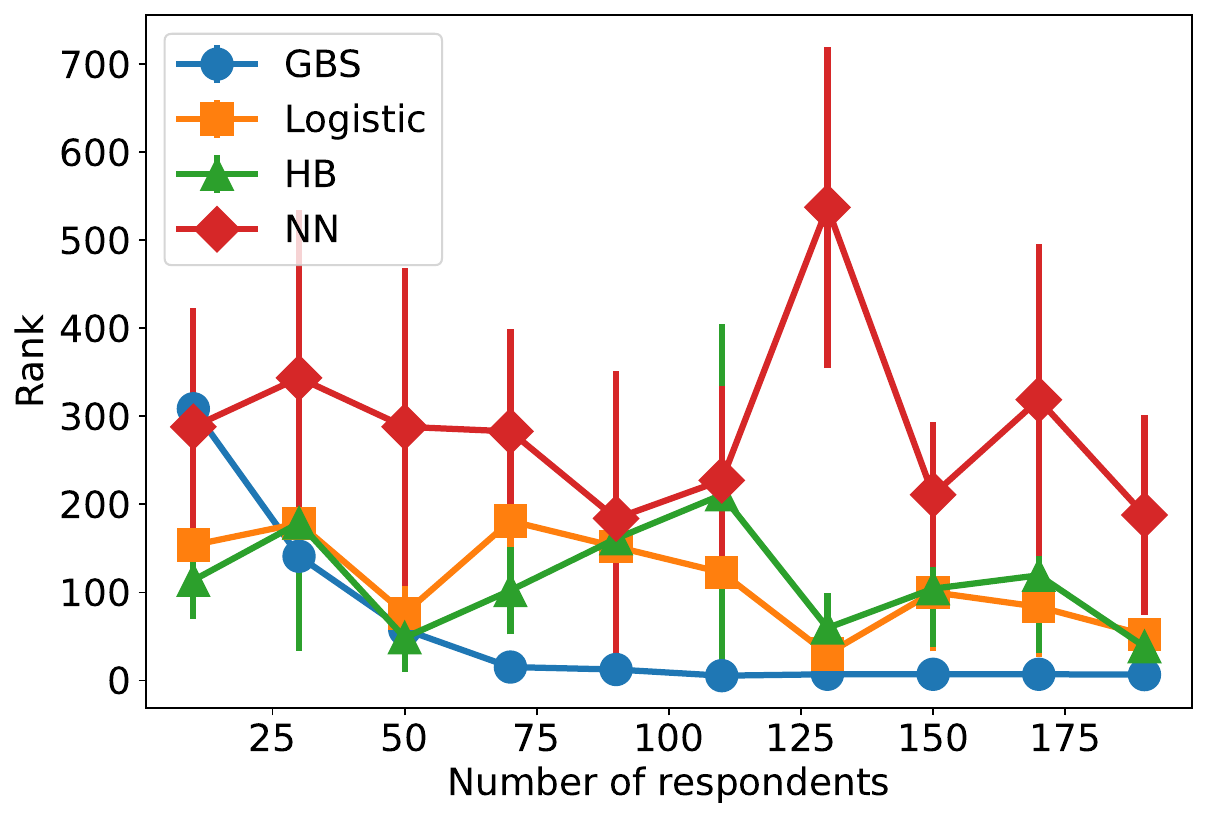}
        \caption{Type 3}
        \label{fig:rt3k10}
        \end{subfigure}
  \caption{The utility (Top row, higher is better) and the ranking (Bottom row, lower is better) for the selected products with $K=10$ features. GBS has similar performance with correctly specified choice utility models (Type 1), and identifies better product when the true utility is unknown (Type 2,3). }
  \label{fig:uk10}
\end{figure}

\vspace{2mm}
\Cref{fig:uk10} shows the test utility and ranking of the estimated optimal product across a different number of respondents for a product with  $K=10$ features. When the utility function is correctly specified (Type 1), Logistic and HB reach the highest utility with a small number of respondents. However, when the utility model is misspecified (Type 2, 3), the performance of Logistic and HB with linearity assumption significantly deteriorates. For Type 2 utility, the ranking drops with an increasing number of respondents when the driving factor for the estimates shifts from variance to bias. NN assumes a flexible nonparametric utility function, but when the sample size is small, its performance is dominated by the variances in the data. \Cref{fig:nnk10} in the \Cref{app:add} shows the results of NN with additional respondents. For the utility function without interactions, it needs 500 respondents to select the optimal product, and for relatively complex Type 2 and 3 utility, the number of respondents needed is 4000 and 2000, respectively. The weak data efficiency increases the experimental cost and may be infeasible. Moreover, finding the optimal product with a trained NN utility function needs to explore all possible products as inputs, which becomes challenging when the feature dimension is high. In comparison, GBS identifies the optimal products across all utility types with less than 100 respondents. For the linear utility, when the respondent number is small,  the correctly specified models outperform GBS, but the performance gap diminishes quickly when the respondent number increases to around 70. For nonlinear utilities, GBS outperforms the baseline models and achieves the global optimum. 

\begin{figure}
  \centering
  \begin{minipage}{0.75\textwidth}
  \begin{subfigure}{.33\textwidth}
  \centering
  \includegraphics[width=.95\linewidth]{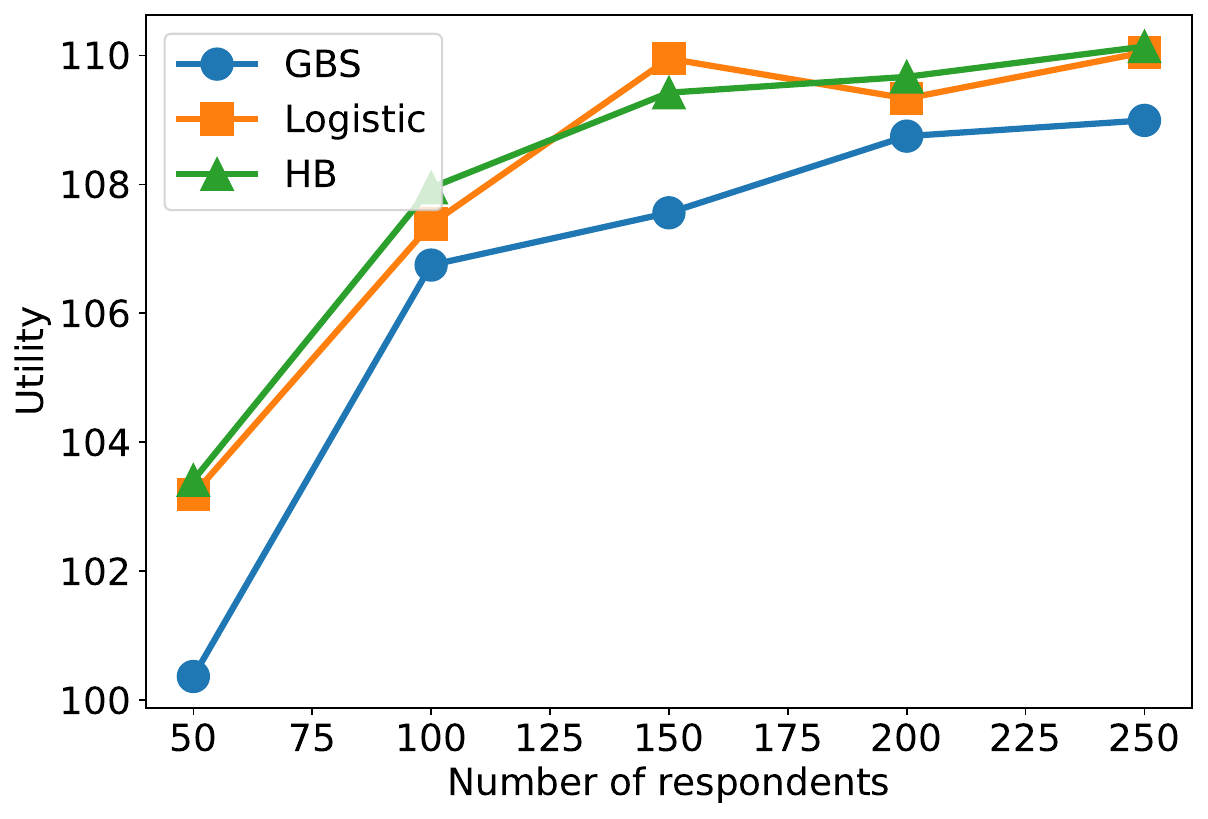}
  \caption{Type 1}
  \label{fig:rt1k100}
  \end{subfigure}%
  \begin{subfigure}{.33\textwidth}
  \centering
  \includegraphics[width=.95\linewidth]{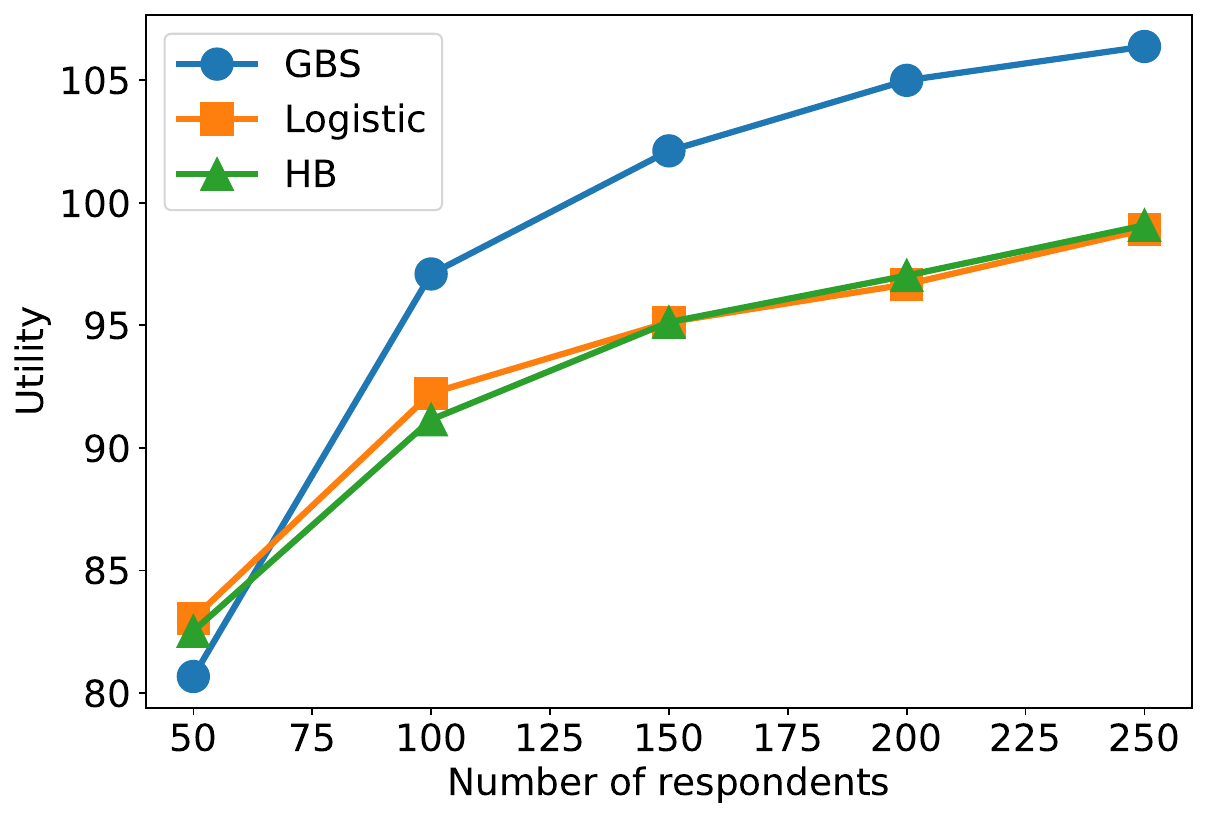}
  \caption{Type 2}
  \label{fig:rt2k100}
  \end{subfigure}
  \begin{subfigure}{.33\textwidth}
    \centering
    \includegraphics[width=.95\linewidth]{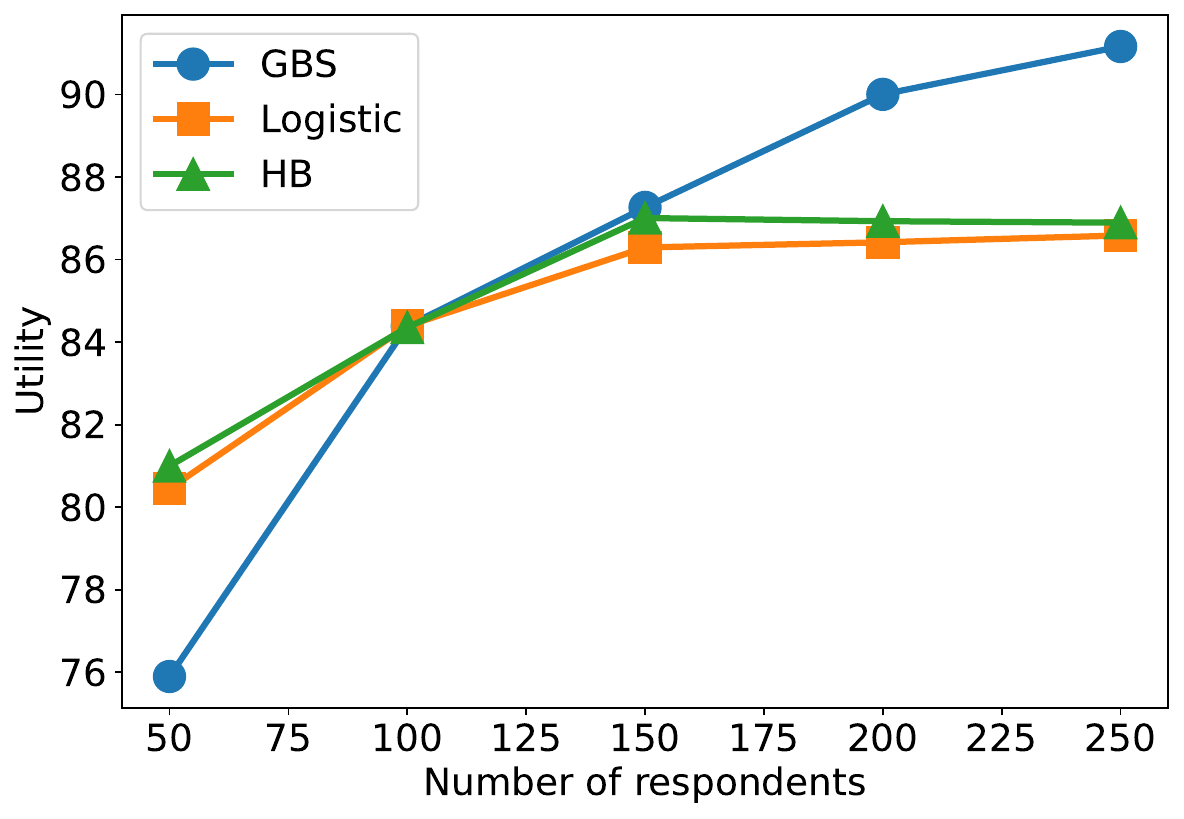}
    \caption{Type 3}
    \label{fig:rt3k100}
    \end{subfigure}
  \caption{The utility (higher is better) with $K=100$ features.}
  \label{fig:uk100}
  \end{minipage}
  \begin{minipage}{0.24\textwidth}
    \centering
  \includegraphics[width=\textwidth]{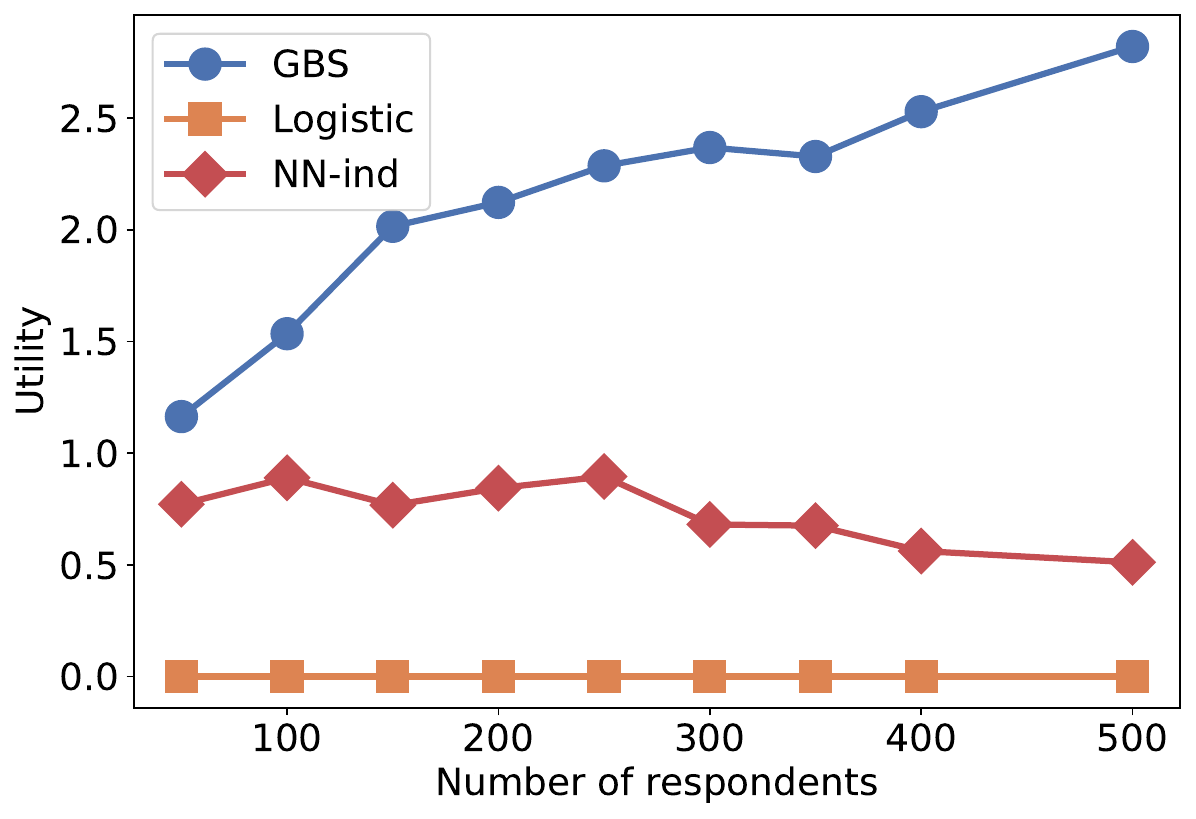}
  \caption{Personalized Product.}
  \label{fig:ind}
  \end{minipage}
\end{figure}

\Cref{fig:uk100} shows the results for the single product design with $K=100$ features. Finding the optimal product with a trained NN needs to compare the predicted utility of $2^{100}$ items, which is infeasible to compute, so we drop it from the baselines. Similarly, we drop the ranking metric that needs to evaluate all product combinations. The performance pattern is similar to $K=10$. GBS can identify the optimal product from around $10^{30}$ choices in less than half a minute. It is flexible with the underlying utility function and is efficient with data size, which makes it practically applicable.

\Cref{fig:uk10,fig:uk100} illustrates a trade-off between specification and estimation \citep{susan2022active}. 
A complex utility model might capture a broader spectrum of choice behaviors and hence has a small specification error; it could also introduce big estimation errors when limited to a small data set. Instead of estimating a full choice model, GBS directly uses the choice data for product design, bypassing this trade-off. GBS is flexible with the underlying utility function and is efficient with data size, which makes it practically applicable.  ~\looseness=-1

Next, we study the personalized product design with choice experiments. The data is generated the same as before, except the individual preferences follow a mixture distribution $W_i \sim 0.5\cN(\mu_1,\Sigma) + 0.5\cN(\mu_2,\Sigma)$ where $\mu_1$ is positive on the first half of elements and negative on the others, and $\mu_2$ is opposite. The mixture distribution reflects the preference heterogeneity in the population. We assume the observed covariates $X_i = \exp(W_i)$ of individual $i$ and take the utility as the nonlinear Type 2. We modify the neural network utility model as $\hat{V}(Z, X_i) = Z^\top f_\gamma(X_i)$ (denoted as NN-ind). The utility form gives an estimated optimal product for individual $i$ as $\mathbf{1}[f_\gamma(X_i)>0]$ without the need to evaluate all possible products' utility for each individual.

\Cref{fig:ind} shows the selected products' utility averaged on a hold-out test set. Since Logistic estimates a single optimal product, the estimate is a  product with zero utility due to the symmetry of the population. NN-ind adapts to individual heterogeneity and has utility higher than Logistic. However, it is subject to model misspecification due to the last linear layer and may have low data efficiency as in the case of the single optimal product. GBS reaches the highest utility, and the performance improves with more respondents joining the experiments. 

\section{Discussions}

This paper bridges the domains of gradient-based machine learning and discrete choice experiments. The proposed GBS is flexible with the underlying form of choice utility, is data-efficient by an adaptive design, is scalable to high-dimensional features, and is applicable to uniform or personalized product designs. 

However, there is no free lunch. GBS does not estimate a choice model. It does not provide a full rank of all the possible products. To explain people's preferences, GBS may need explainable AI techniques such as saliency map \citep{adebayo2018sanity} rather than estimating the preferences as a part of model parameters. Nevertheless, GBS provides a flexible optimization framework. Except for maximizing the market share, the objective may incorporate the costs and prices of a product to maximize the profit. If the product design is under a budget constraint, the constrained optimization might be considered using methods such as projected gradient. 

From a manager's view, a company often needs to design a product line consisting of several products \citep{balakrishnan2004development,belloni2008optimizing}. One way to apply GBS for this task is by a separate approach, where the population is clustered into segments, and a single best product is determined for each segment \citep{Paetz2021}. It is also feasible to model a product line as several binary vectors and apply GBS to design a product line jointly. ~\looseness=-1

GBS builds on the inference of discrete latent variables in machine learning \citep{gu2015muprop,jang2016categorical,tucker2017rebar,kool2019buy,disarm2020,kunes2023gradient}. If a gradient estimator contrasts several values of the objective function, it can potentially be used for the adaptive question design. Combining recent discrete optimization techniques with discrete choice experiments is an interesting future direction.

\FloatBarrier
\bibliographystyle{apalike}
\bibliography{ref}

\appendix
\section{Proof}
\label{app:theorem}

This section provides the proof of  \Cref{lemma:grad,lemma:zero,lemma3}. \\

\begin{proof}[Proof of \Cref{lemma:grad}] The optimization objective is
  \bas{
    V(\phi) = \bE_{Z \sim p(Z;\sigma(\phi))}\bE[Y_i(Z, Z_0)\g Z],
  }
  where $p(Z;\sigma(\phi)) = \prod_{k=1}^K\text{Bern}(z_k; \sigma(\phi_k))$, $Z = (z_1, \cdots, z_K)$. The k-th element for the gradient of $V(\phi)$ is
  \ba{
\nabla_{\phi_k}V(\phi) &=\bE_{\zv_{\backslash k}\sim\prod_{\nu\neq k}{\text{Bern}}(z_\nu;\sigma(\phi_\nu))} \{ \nabla_{\phi_k} \bE_{z_k\sim{\text{Bern}}(\sigma(\phi_k))}\bE[Y_i(Z, Z_0)\g Z]\}
\label{eq:pf1}
}

Denote $f(z_k) = \bE[Y_i(Z, Z_0)\g Z]$, we have
\ba{
  &\nabla_{\phi_k} \bE_{z_k\sim{\text{Bern}}(\sigma(\phi_k))}[f(z_k)]  \notag \\
  =& \sigma(\phi_k)\sigma(-\phi_k)[f(1)-f(0)]  \notag\\
  =& \bE_{u\sim \text{Unif}(0,1)}[f(\mathbf{1}{[u<\sigma(\phi)]})(1-2u)]  \notag\\
  =& \bE_{ u\sim\text{Unif}(0,1)}[f(\mathbf{1}{[u< \sigma(\phi) ]})(1/2-u)] + \bE_{ \tilde u\sim\text{Unif}(0,1)}[f(\mathbf{1}{[\tilde u< \sigma(\phi) ]})(1/2-\tilde u)]  \notag\\
  =& \bE_{u\sim\text{Unif}(0,1)}\left[\big(f(\mathbf{1}{[u>\sigma(-\phi)]}) - f(\mathbf{1}{[u<\sigma(\phi)]} )\big) (u-1/2) \right ].  
  \label{eq:pf2}
}
The first two equations  can be straightfowardly evaluated since it is an expectation with a scalar variable. The third equation is applying antithetic sampling with $\tilde u=1-u$. The last equation is summing up the two expectations. Plugging \Cref{eq:pf2} into \Cref{eq:pf1} gives

\bas{
  &\nabla_{\phi_k}V(\phi)  \\
  =&\bE_{\zv_{\backslash k}\sim\prod_{\nu\neq k}{\text{Bern}}(z_\nu;\sigma(\phi_\nu))}\big \{\bE_{u_k\sim {\text{Unif}}(0,1)}\Big[ \big( \bE[Y_i(Z, Z_0)\g Z=(\zv_{\backslash k}, z_k=\mathbf{1}{[u_k>\sigma(-\phi_k)]})] -  \\
  & \bE[Y_i(Z, Z_0)\g Z=(\zv_{\backslash k}, z_k=\mathbf{1}{[u_k<\sigma(\phi_k)]})] \big) (u_k-\frac12)\Big] \\
  =&  \bE_{\uv\sim\prod_{k=1}^K{\text{Unif}}(0,1)}\Big[  \big(  \bE[Y_i(Z, Z_0)\g Z=\mathbf{1}{[\uv>\sigma(-\phiv)]})] -  \bE[Y_i(Z, Z_0)\g Z=\mathbf{1}{[\uv < \sigma(\phiv)]})] \big) (u_k-\frac12)\Big]
}
Therefore,
\ba{
  \nabla_\phi V(\phi) = \bE_{u \sim \prod_{k=1}^K\text{Unif}(0,1)} \left[\bE[(Y_i(Z_1(u),Z_0) - Y_i(Z_2(u),Z_0))(u-\frac12) \g u]\right],
  }
where $Z_1(u) = \mathbf{1}[u > \sigma(-\phi)], Z_2(u) = \mathbf{1}[u < \sigma(\phi)].$
\end{proof}

\begin{proof}[Proof of \Cref{lemma:zero}] 
Denote $S = Y(Z_1,Z_0) - Y(Z_2,Z_0)$, $T=2Y(Z_1, Z_2) - 1$. We have
\bas{
p(S=1) =& p(Y(Z_1,Z_0)=1, Y(Z_2,Z_0)=0)  \\
=& \frac{e^{U(Z_1)}}{e^{U(Z_1)}+ e^{U(Z_0)}} \cdot \frac{e^{U(Z_0)}}{e^{U(Z_2)}+ e^{U(Z_0)}} 
}
and
\bas{
  &p(S=1 \g Y(Z_1,Z_0) \neq Y(Z_2,Z_0)) \\
  =& \frac{p(Y(Z_1,Z_0)=1, Y(Z_2,Z_0)=0)}{p(Y(Z_1,Z_0)=1, Y(Z_2,Z_0)=0)+p(Y(Z_1,Z_0)=0, Y(Z_2,Z_0)=1)} \\
  =& \frac{e^{U(Z_1)+U(Z_0)}}{e^{U(Z_1)+U(Z_0)} + e^{U(Z_2)+U(Z_0)}} \\
  =& \frac{e^{U(Z_1)}}{e^{U(Z_1)} + e^{U(Z_2)}} \\
  =& p(Y(Z_1,Z_2)=1) \\
  =& p(T=1).
}
Similarly, we have $p(S=-1 \g Y(Z_1,Z_0) \neq Y(Z_2,Z_0)) = p(T=-1)$. Therefore, $S \g Y(Z_1,Z_0) \neq Y(Z_2,Z_0) \stackrel{d}{=} T$.
\end{proof}

\begin{algorithm}[t]
  \small{
  \SetKwData{Left}{left}\SetKwData{This}{this}\SetKwData{Up}{up}
  \SetKwFunction{Union}{Union}\SetKwFunction{FindCompress}{FindCompress}
  \SetKwInOut{Input}{input}\SetKwInOut{Output}{output}
  \Input{Individual covariates $\{X_i\}$, policy function $g(\cdot; \theta)$, number of features $K$, number of questions per respondent $n_q$, stepsize $\eta$
  }
  
  Initialize the policy parameters $\theta$ randomly.
  
  \While{not converged}{

   Sample a random individual $i$ from the population.

   \For{$j = 1,2 \cdots n_q$}{
    Sample $u \sim \prod_{k=1}^K\text{Unif}(0,1)$

    Generate a pair of products $Z_1(u) = \mathbf{1}[u > \sigma(-\phi)], Z_2(u) = \mathbf{1}[u < \sigma(\phi)]$
    
    Record the respondent's choice $Y_i(Z_1(u), Z_2(u))$

    Compute the gradient estimate $g_{\text{GBS}}(\theta)$ by \Cref{eq:grad4}

    Update $\theta \leftarrow \theta + \eta g_{\text{GBS}}(\theta)$
   }
  }
  \caption{Gradient-based Survey for Policy Learning}
  \label{alg:gbs-ind}}
\end{algorithm}

\begin{proof}[Proof of \Cref{lemma3}]
  Denote the event $\mathcal{A}_i = \{Y_i(Z_1,Z_0) \neq Y_i(Z_2,Z_0)\}$, we have
  \ba{
    & \bE_{u \sim \prod_{k=1}^K\text{Unif}(0,1)} \left[\bE\Big[(Y_i(Z_1(u),Z_0) - Y_i(Z_2(u),Z_0))(u-\frac12) \g u\Big]\right] \\ =& \bE_{u \sim \prod_{k=1}^K\text{Unif}(0,1)} \left[\bE\Big[(Y_i(Z_1(u),Z_0) - Y_i(Z_2(u),Z_0))(u-\frac12) \g u,\mathcal{A}_i\Big]p(\mathcal{A}_i) \right]  \\
    =& \bE_{u \sim \prod_{k=1}^K\text{Unif}(0,1)} \left[\bE\Big[(2Y_i(Z_1(u), Z_2(u)) - 1) (u-\frac12) p(\mathcal{A}_i) \g u\Big]\right].
    \label{eq:grad4}
  } 
  The first equality is by the law of total expectation; the second is by \Cref{lemma:zero} and the law of unconscious statistician (LOTUS).  The Monte Carlo estimation of the gradient in \Cref{eq:grad4} is $\tilde{g} = (2Y_i(Z_1(u), Z_2(u)) - 1) (u-\frac12) p(\mathcal{A}_i)$, $u \sim \prod_{k=1}^K\text{Unif}(0,1)$.
\end{proof}

\section{Additional Results}
\label{app:add}

\begin{figure}[ht]
  \centering
    \begin{subfigure}{.33\textwidth}
  \centering
  \includegraphics[width=.95\linewidth]{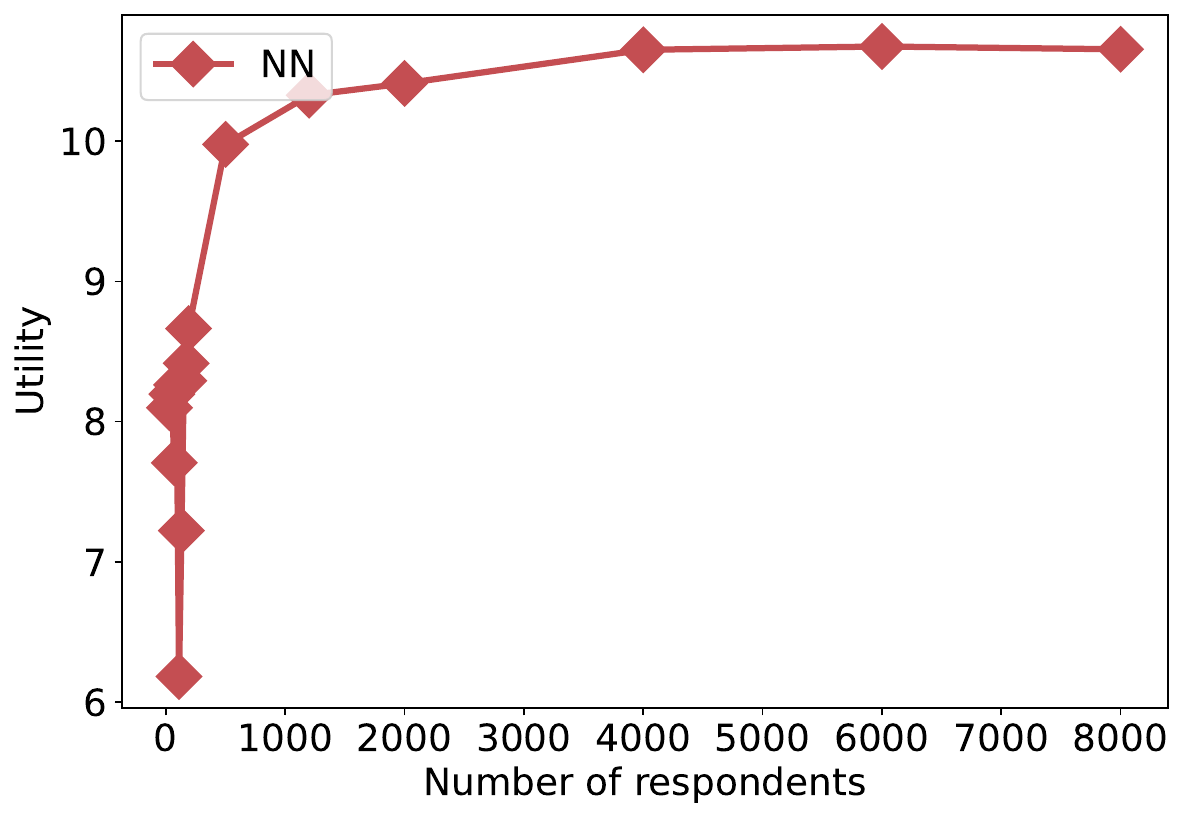}
  \end{subfigure}%
  \begin{subfigure}{.33\textwidth}
  \centering
  \includegraphics[width=.95\linewidth]{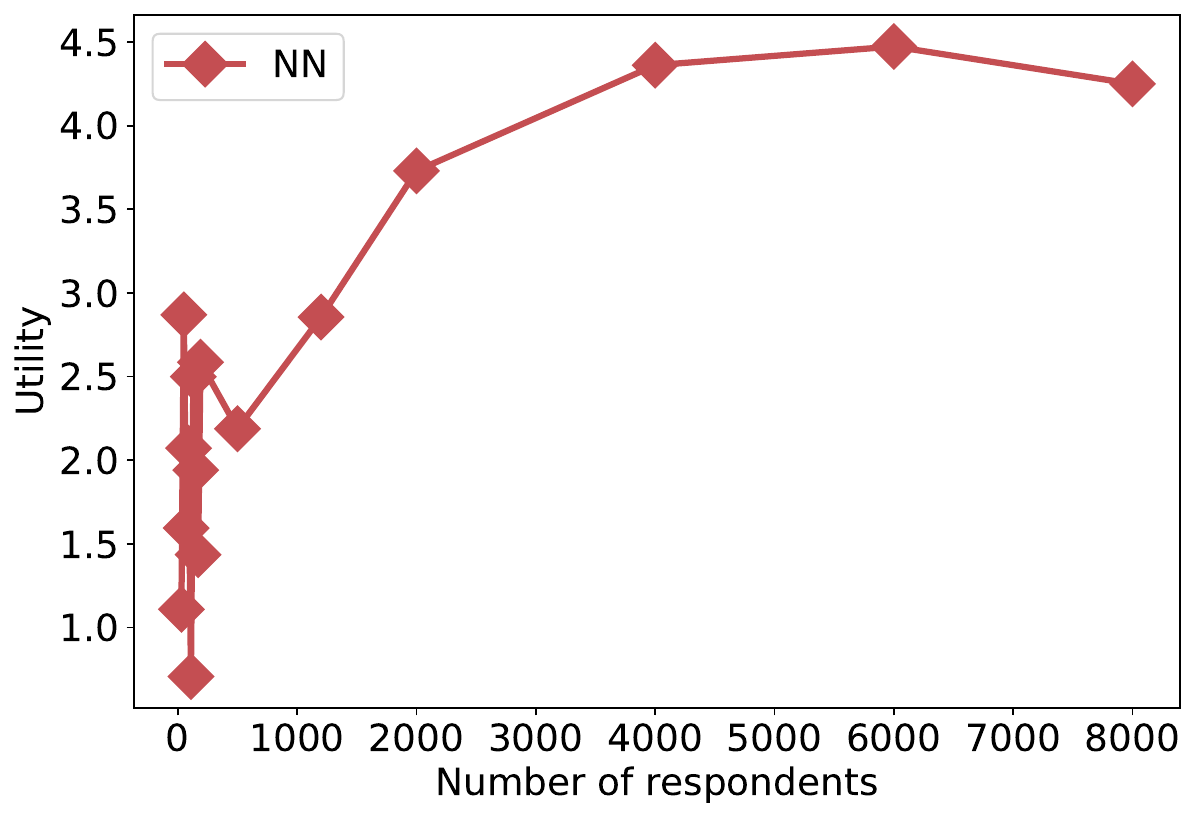}
  \end{subfigure}
  \begin{subfigure}{.33\textwidth}
    \centering
    \includegraphics[width=.95\linewidth]{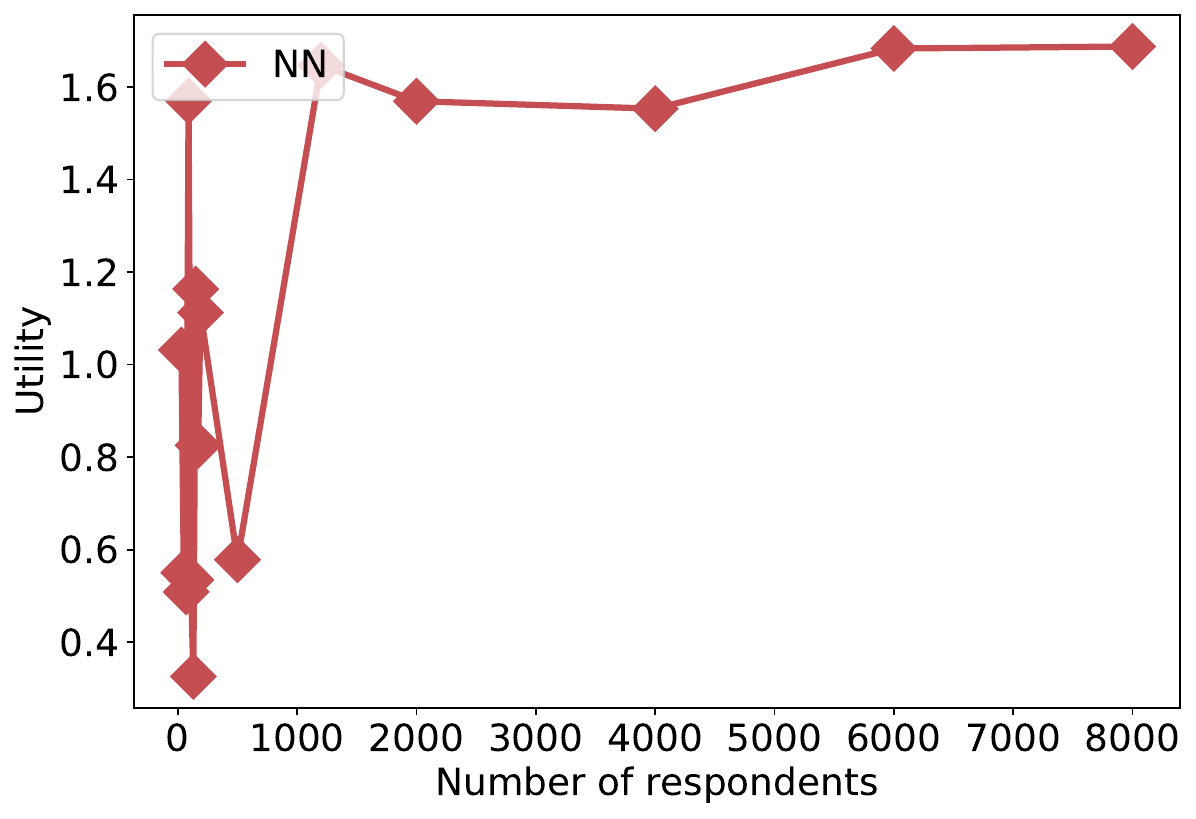}
    \end{subfigure}
  \begin{subfigure}{.33\textwidth}
  \centering
  \includegraphics[width=.95\linewidth]{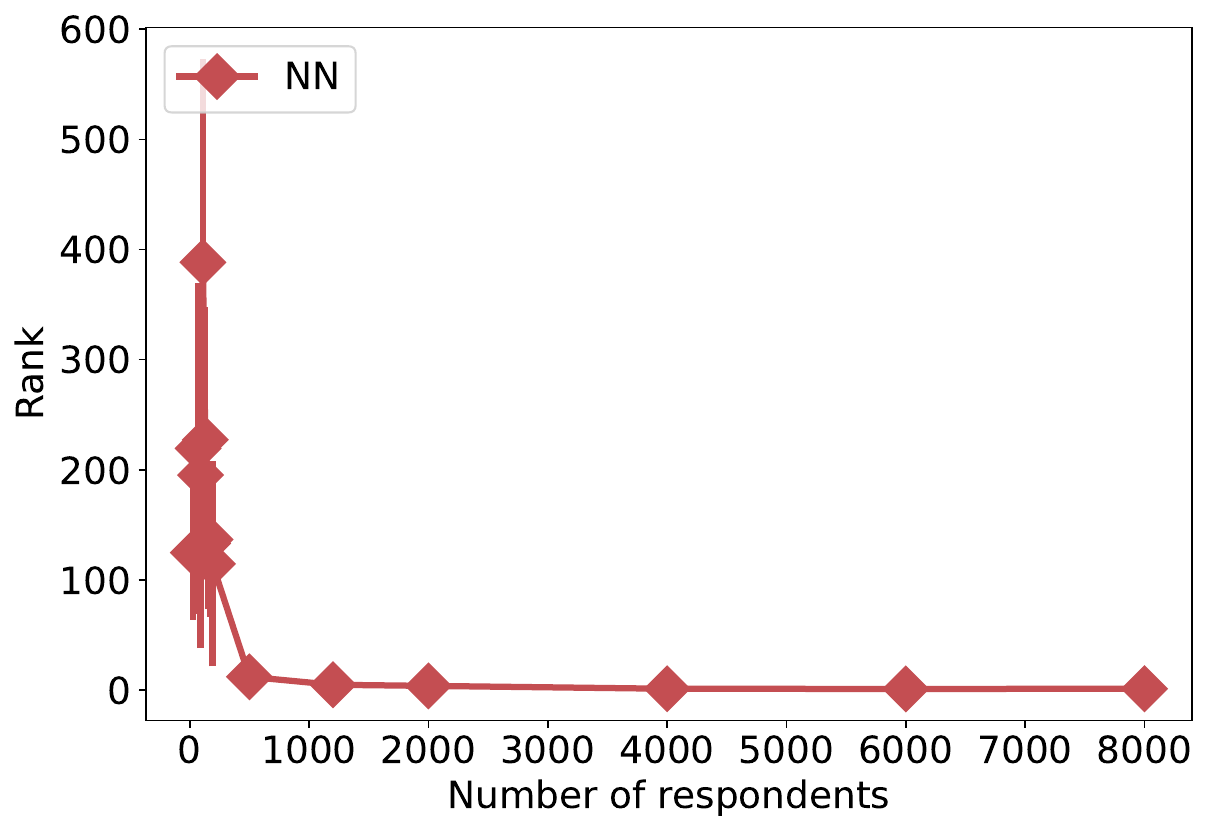}
  \caption{Type 1}
  \end{subfigure}%
  \begin{subfigure}{.33\textwidth}
  \centering
  \includegraphics[width=.95\linewidth]{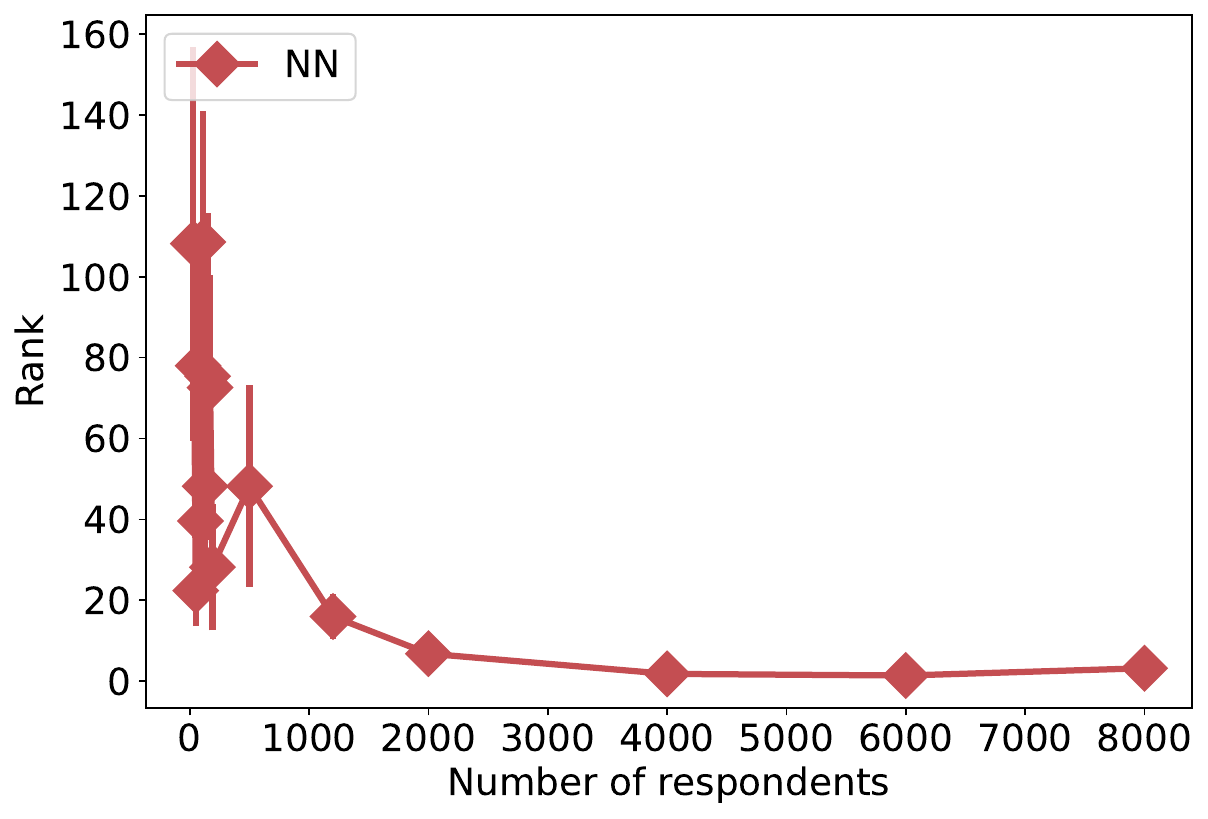}
  \caption{Type 2}
  \end{subfigure}
  \begin{subfigure}{.33\textwidth}
    \centering
    \includegraphics[width=.95\linewidth]{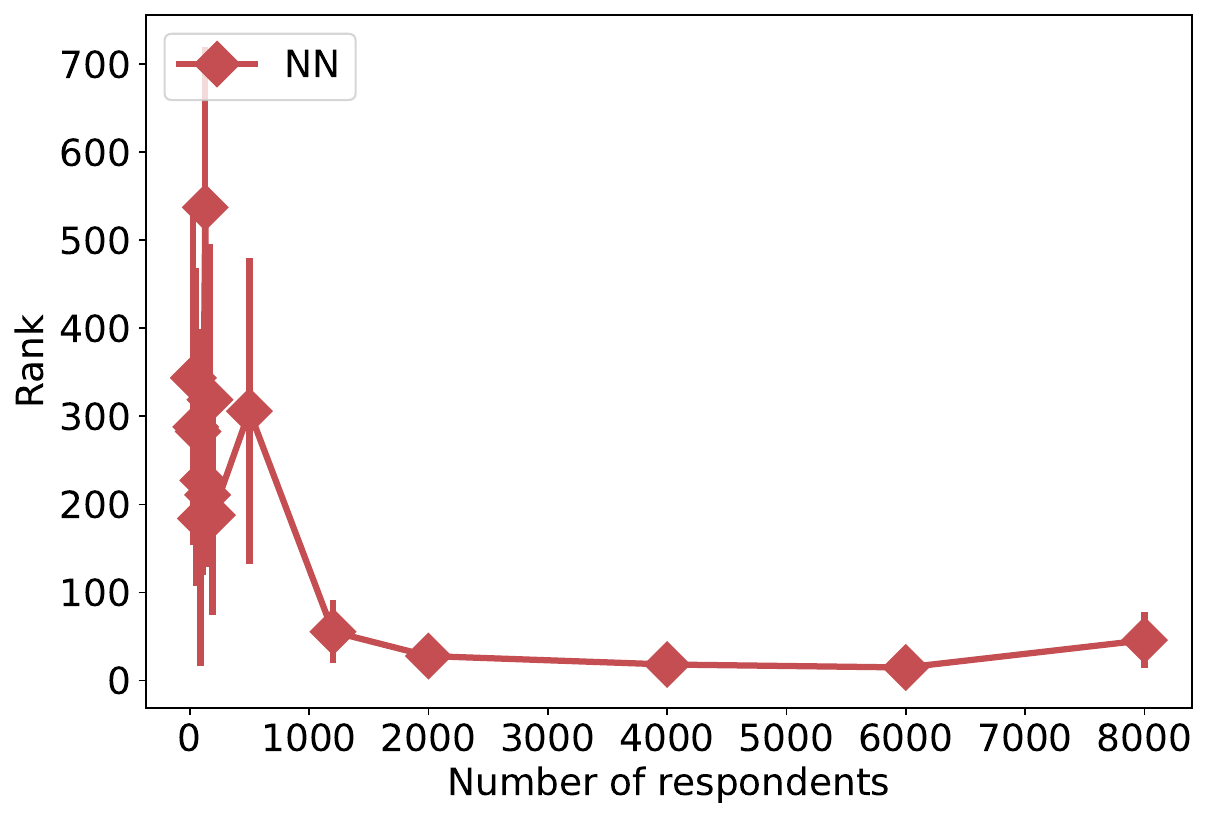}
    \caption{Type 3}
    \end{subfigure}
  \caption{The utility (Top row) and the ranking (Bottom row) for the NN baseline with a large number of respondents.}
  \label{fig:nnk10}
\end{figure}

This section contains the customized product  design algorithm and additional results for the empirical study in \Cref{sec:exp}. \Cref{fig:nnk10} shows the utility and ranking for the NN baseline with a large number of respondents.

\end{document}